\documentclass[letterpaper, 10 pt, conference]{ieeeconf}  

\IEEEoverridecommandlockouts                              
\usepackage{amssymb}
\usepackage{amsmath}
\usepackage{bm}
\usepackage{graphics}
\usepackage{epsfig}
\usepackage{multirow}
\usepackage{nicefrac}
\usepackage{cases}          
\usepackage{subcaption}
\usepackage{mathtools}
\usepackage{xcolor}
\usepackage{algorithm}
\usepackage{algorithmicx}
\usepackage{algpseudocode}
\usepackage{balance}

\newcommand{\mat}[0]{\begin{bmatrix}}
\newcommand{\mate}[0]{\end{bmatrix}}
\newcommand{\va}{\mathbf{a}}
\newcommand{\vb}{\mathbf{b}}

\newcommand{\ve}{\mathbf{e}}

\newcommand{\vp}{\mathbf{p}}
\newcommand{\vq}{\mathbf{q}}

\newcommand{\vs}{\mathbf{s}}

\newcommand{\vu}{\mathbf{u}}
\newcommand{\vv}{\mathbf{v}}

\newcommand{\vx}{\mathbf{x}}
\newcommand{\vy}{\mathbf{y}}

\newcommand{\vo}{\mathbf{o}}

\newcommand{\cC}{\mathcal{C}}

\newcommand{\cH}{\mathcal{H}}

\newcommand{\cJ}{\mathcal{J}}

\newcommand{\cO}{\mathcal{O}}
\newcommand{\cP}{\mathcal{P}}

\newcommand{\cT}{\mathcal{T}}
\newcommand{\cU}{\mathcal{U}}

\newcommand{\cY}{\mathcal{Y}}

\newcommand{\R}{\mathbb{R}}

\newcommand\norm[1]{\left\|#1\right\|}              

\definecolor{wb}{rgb}{0, 0.5, 0}
\definecolor{removed}{rgb}{0.9, 0.9, 0.9}

\newcommand{\wbr}[2]{{\color{wb}#2}}
\newcommand{\rebuttalsub}[2]{{\color{black}#2}}

\usepackage{siunitx}

\definecolor{blue}{rgb}{0.0, 0.0, 1.0}
\newcommand{\newcont}[1]{{\color{black}{\sc}#1}}

\title{\LARGE \bf
Active Classification
of Moving Targets with Learned Control Policies
}

\author{\'{A}lvaro Serra-G\'{o}mez$^{1}$, Eduardo Montijano$^{2}$, Wendelin Böhmer$^{3}$, Javier Alonso-Mora$^{1}$ 
\thanks{This work is supported by the U.S. Office of Naval Research Global (ONRG) NICOP-grant N62909-19-1-2027}
\thanks{$^1$Department of Cognitive Robotics, Delft University of Technology; {\tt\small $\{$a.serragomez; j.alonsomora$\}$@tudelft.nl}}%
\thanks{$^2$Department of Informatics and Systems Engineering, Universidad de Zaragoza; {\tt\small emonti@unizar.es}}%
\thanks{$^3$Department of Software Technology, Delft University of Technology; {\tt\small j.w.bohmer@tudelft.nl}}
}

\usepackage{graphicx}
\usepackage{dblfloatfix}

\begin{document}

\maketitle
\thispagestyle{empty}
\pagestyle{empty}

\begin{abstract}

In this paper, we consider the problem where a drone has to collect semantic information to classify multiple moving targets. In particular, we address the challenge of computing control inputs that move the drone to informative viewpoints, position and orientation, when the information is extracted using a ``black-box'' classifier, e.g., a deep learning neural network. These algorithms typically lack of analytical relationships between the viewpoints and their associated outputs, preventing their use in information-gathering schemes.  
To fill this gap, we propose a novel attention-based architecture, trained via Reinforcement Learning (RL), that outputs the next viewpoint for the drone favoring the acquisition of evidence from as many unclassified targets as possible while reasoning about their movement, orientation, and occlusions. Then, we use a low-level MPC controller to move the drone to the desired viewpoint taking into account its actual dynamics. 
We show that our approach not only outperforms a variety of baselines but also generalizes to scenarios unseen during training. Additionally, we show that the network scales to large numbers of targets and generalizes well to different movement dynamics of the targets.

\end{abstract}


\section{Introduction}\label{sec:introduction}
In surveillance and tracking applications an autonomous drone may be tasked with collecting relevant information from multiple targets, e.g., recognize people with blue eyes.
Recent deep learning approaches show excellent results at detecting and categorizing single and multiple elements with images~\cite{Redmon2016YouOL} or LiDAR~\cite{Alonso-21}. However, these methods are generally 
not enough for active classification with a mobile drone, which also requires planning of the drone's movement and reasoning over the targets' future behavior.

The use of deep learning perception algorithms for information gathering comes with its own challenges. On the one hand, the ``black-box'' nature of these algorithms makes it difficult to 
determine the position that would yield the most informative data for classification.
On the other hand, the drone also needs to reason about the targets' movement and orientation, as well as the possible occlusions among them, 
to plan a trajectory that will reveal the most information.

To overcome these issues, the main contribution of this paper is a complete solution to the problem of active classification of multiple moving targets.
Differently to previous approaches, our framework can handle \emph{dynamic} targets without requiring an explicit observation model, e.g., using a black-box classifier.
Our solution leverages Deep Reinforcement Learning (DRL) to train a control policy that recommends informative viewpoints using the relative position of the targets and their current class probability estimations. 
\rebuttalsub{The policy is encoded by a novel attention-based architecture able to implicitly reason about how 
different amounts of targets affect the quality of each other's observations\newcont{, and the weight of each in its viewpoint recommendation}.
The architecture is also scalable, performing successfully in scenarios with larger amounts of targets than those seen during training.}{We also propose a novel attention-based permutation-invariant architecture for the DRL policy that generalizes to more targets that move differently from those seen during training.}

\begin{figure}[t!]
    \centering
    \includegraphics[width=0.90\linewidth]{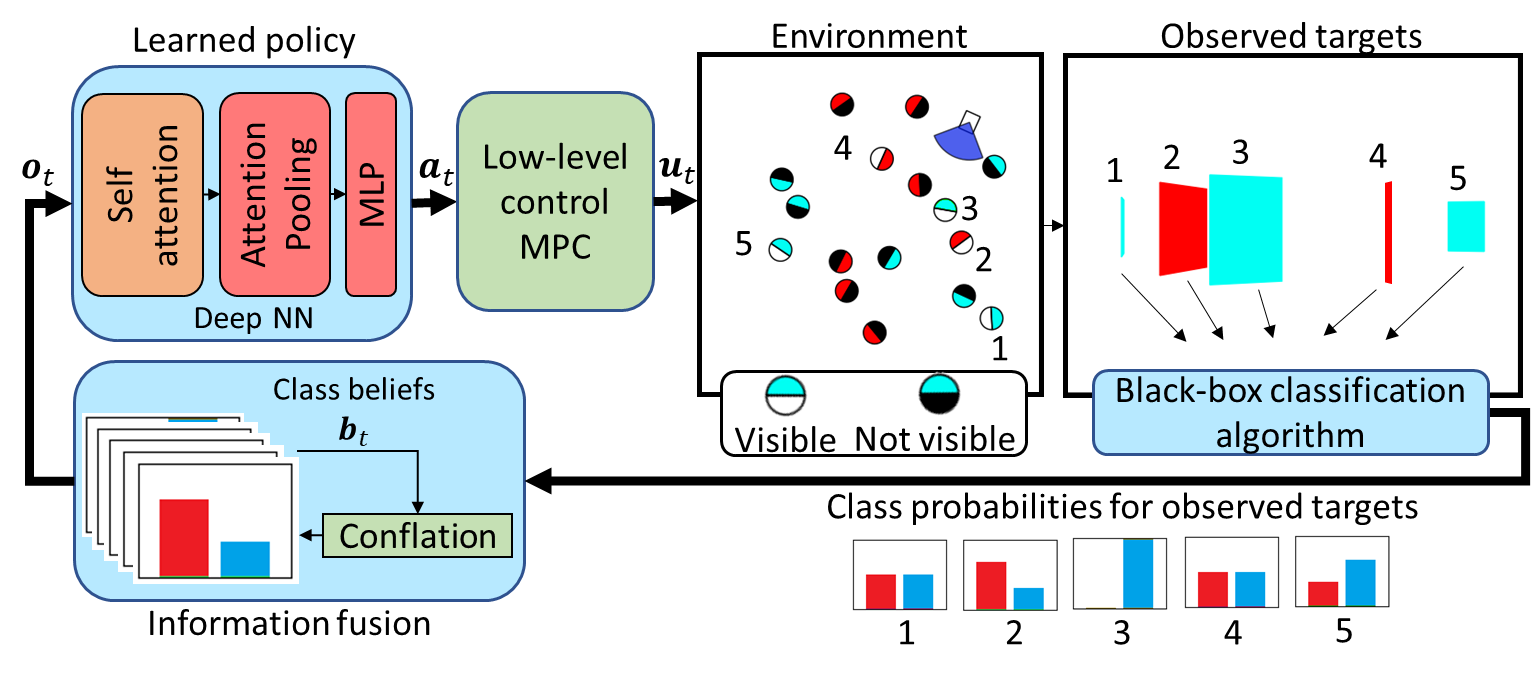}
    \caption{\footnotesize{Overview of our method for active classification. The objective is to classify all targets into different classes, red and blue in the example.  
    The method has three parts.
    First, observable targets, targets 1 to 5, represented with white back in the figure, are detected by the onboard sensor. We assume the existence of a classification algorithm, e.g., a CNN, able to provide probability estimates for each target (bars at the bottom right). Then, we use an efficient fusion mechanism, conflation, to obtain the beliefs, $\vb_t,$ over the classes for each target.
    Finally, this information, together with the observed probabilities and relative locations, $\vo_t$, is fed to a control algorithm that computes the recommended viewpoint, $\va_t$, using a novel deep learning architecture trained using Reinforcement Learning. The recommended viewpoint is tracked with a low-level controller, which outputs the control input $\vu_t$ sent to the robot.
    }}
    \label{fig:overview}
\end{figure}

Moreover, to simplify the training, the policy is abstracted from the low-level dynamics of the drone, which are instead considered inside a low-level MPC controller at test time.
Finally, the estimations are updated with an efficient information fusion method, conflation, suited to be used with black-box perception algorithms. A full overview of the method is shown in Fig.~\ref{fig:overview}.
Experiments under different conditions show that our approach outperforms a variety of baselines and is robust to scenarios unseen during training.


\section{Related Works}

Our contributions build upon recent work in multi-view active classification and learning for motion planning in dynamic environments.


\subsection{Multi-view active classification}
The problem of active classification is typically solved by pre-defining a set of viewpoints through which trajectories are planned to gather sufficient information to solve an active classification task. 
One-step greedy planners 
for active object classification~\cite{Patten2016ViewpointEF} select object-dependent viewpoints based on class uncertainty and observation occlusions. 
Some non-myopic methods~\cite{activeWeedClassification} account for the cost of movement and information gain between viewpoints to solve the problem of active classification.
Others, e.g.,~\cite{Natansov2014}, formulate the problem as a partially observable Markov Decision Process (POMDP) and plans a path over viewpoints while accounting for measurement costs, occlusions and possible misclassifications.
With a similar formulation,~\cite{Patten2018} computes plans using a variation of Monte Carlo tree search. However, these methods generally require access to an observation model to estimate viewpoint usefulness a priori. 

Learning methods are non-myopic and allow to use black-box models. Recent works leverage the use of DRL for static multi-target pose estimation and active classification by learning a policy that moves the camera towards viewpoints 
that reduce observation uncertainty~\cite{active6D} or maximize information gain to enhance object classification~\cite{efficientMultiview}.
Nevertheless, these methods assume that targets are static, which makes them unsuitable for active classification with moving targets. Although some works consider dynamic targets, they are often limited to a pre-defined closed environment~\cite{activePerceptionAutonomousObservation} or, as its the case for aerial videography methods~\cite{ALCANTARA2021103778, detectionAware}, focus on target information visibility which requires prior knowledge on where this information is visible from.

\subsection{Learning for motion planning in dynamic environments}
Learning approaches in motion planning tasks have the potential to encode and identify patterns 
in complex motion planning tasks. Recent advances in Deep Representation Learning show promise in learning latent representations of the state space, capturing the underlying structure and symmetries in dynamic environments~\cite{GeometricDL}. Recent works in learning-based motion-planning policies rely on Convolutional Neural Networks (CNN)~\cite{DenseCAvoid,cui2021} for visual-based navigation, or encode the state of multiple 
static/dynamic elements in the environment using Graph Convolutional Networks (GCN)~\cite{Li2019,Chenyuying} or Long-Short Term Memory layers (LSTMs)~\cite{britogompc,meverett2021}. 

Yet these strategies require a priori knowledge on the priority order of the elements in the encoded sequence, or the structure of the encoded graph~\cite{kurin2021my}. Instead, DeepSets~\cite{NIPS2017_f22e4747} and self-attention based architectures~\cite{Chen2019,NIPS2017_3f5ee243} are permutation invariant and enable encoding element sets without making further assumptions on their structure. 
Most related to our approach,~\cite{Hsu2021} uses a combination of Self-attention~\cite{NIPS2017_3f5ee243} and DeepSets~\cite{NIPS2017_f22e4747} to learn a policy 
{that coordinates} a multi-robot system to track a set of dynamic targets. 
{However}, DeepSets aggregate all targets' information by assigning equal weights to them. In contrast, we leverage the use of self-attention~\cite{NIPS2017_3f5ee243} and attention-based permutation-invariant set-function approximators~\cite{lee2019set} to effectively encode the dynamic environment, learning the importance of each target in our future decision.

\section{Problem formulation}\label{sec:problemformulation}
\def\droneid{0} 



Consider a drone with state $\vx_t$ at discrete time $t$, control inputs $\vu_t$ and dynamics $\vx_{t+1} = f(\vx_t,\vu_t)$, obtained for a sampling time of $\tau_l$ seconds. The drone has to collect information from a set 
$\mathcal{J} = \{1,...,M\}$
of $M$ dynamic targets, where $\vy_t^j$ is the state of target $j$ at time step $t$, represented in the drone's reference frame and $\mathcal{Y}_t=\{\vy^j_t\}_{j=1,...,M}$. The targets follow their own dynamics, $\vy^j_{t+1}=g^j(\vy^j_t)$, which are unknown to the drone. 
We assume that the drone flies above the targets, neglecting physical collisions with them.
Nevertheless, the targets can still collide with each other and, more importantly, they can occlude the visibility of others to the drone, depending on where they are. Besides, we do not deal with how to measure and track\wbr{ing}{} the targets' relative information, assuming they are provided by some external perception algorithm, e.g.,~\cite{sung2017}.\newcont{ For the sake of simplicity, in the following we omit the $t$ subscript, except when needed.}


We denote by $\cC = \{1,...,C\}$ the set of classes, such that each target in $\mathcal{J}$ belongs to one class in $\cC$ (e.g., eye color).
The objective of the drone is to classify all the targets. To do that, the drone has a belief vector for each target, \rebuttalsub{$\vb^j_t = \{b_{t}^{j1},b_{t}^{j2},...,b_{t}^{jC}\},$}{$\vb^j = \{b^{j1},b^{j2},...,b^{jC}\},$} where \rebuttalsub{$0\le b_{t}^{jc}\le 1$}{$0\le b^{jc}\le 1$} denotes the belief the drone has of target $j$ belonging to class $c$ at time $t$, and \rebuttalsub{$\sum_{c=1}^C b_{t}^{jc}=1$}{$\sum_{c=1}^C b^{jc}=1$}.
Target $j$ is tagged as classified whenever \rebuttalsub{$\max_{c\in C}b_t^{jc}$}{$\max_{c\in C}b^{jc}$} is above a pre-defined threshold $b_{\max}$. We use the Boolean variable \rebuttalsub{$l^j_t$}{$l^j$}, to specify whether target $j$ is classified or not\newcont{ at timestep $t$}.

To compute the beliefs,
every $\tau_{h} \gg \tau_l$ seconds, the drone is able to make an observation and use a black-box perception algorithm (e.g., a pre-trained CNN classifier) to compute a probability distribution over the class set for each target. We denote 
{that distribution as} \rebuttalsub{$\cP_t = h(\cY_t) = \{\vp^j_t\}_{j=1,...,M}$,
where $\vp_t^j = (p_t^{j1},...,p_t^{jC})$}{$\cP = h(\cY) = \{\vp^j\}_{j=1,...,M}$, where $\vp^j = (p^{j1},...,p^{jC})$} is the probability vector for target $j$, and \rebuttalsub{$p^{jc}_t$}{$p^{jc}$} is the probability of target $j$ belonging to class $c$. For unobserved targets, \rebuttalsub{$\vp_t^j$}{$\vp^j$} is a uniform distribution. 
As it happens with the majority of real CNN classifiers, we assume the drone has no available model to map
how the relative positions relate to the observed probability distributions.
Beliefs are then computed by fusing these measurements,  ${\vb}^j_{t}=\zeta(\vp^j_{1:t})$, where $\zeta$ is the conflation operator~\cite{Hill2011}, a function that models how the beliefs can be computed from the history of classification probabilities given by the black-box sensor~(see Section~\ref{sec:method_belief_updates} for details). 

Under these conditions, the problem considered in the paper is to actuate the drone in such a way that it is able to classify all targets as quickly as possible, i.e., make $l^j_t$ true for all $j$ in the minimum possible value of $t$.
To address it, we formulate a sequential decision-making problem that the drone solves at every time step $t$.
The objective of this problem is to find the actions over a time horizon $T$ that minimize the accumulated entropy of all the targets' beliefs,
\begin{equation} \label{problemformulation}
    \begin{aligned}
    \min_{\vu_{0:T}} & \sum_{t=1}^{T} \sum_{j\in \cJ}
    w_{\mathcal{H}}\mathcal{H}[\vb^j_t] + w_{u} \norm{\mathbf{u}_t}\\
     \text{s.t.} \quad & \vx_{t+1}=f(\vx_t,\vu_t), \quad \vy^j_{t+1}=g^j(\vy^j_t), \quad \forall t\\
        & \cP_t = h(\cY_t), \quad {\vb}^j_{t}=\zeta(\vp^j_{1:t}),\quad \forall t\propto \tau_h/\tau_l\\
        & j \in \cJ, \quad 0 \leq t \leq T\!-\!1
    \end{aligned}
\end{equation}
where $\mathcal H[\vb^j_t]$ denotes the entropy of belief $\vb^j_t$, $w_{\mathcal{H}}$ and $w_{u}$ are scaling weights, and $\propto$ is the proportional sign.

\section{Methodology} \label{sec:method}

The lack of information about $h(\cY_t)$ and $g^j(\vy_t^j)$ hinders the direct solution of problem~\eqref{problemformulation}.
Instead, in the paper we leverage Reinforcement Learning to train a control policy that implicitly learns these quantities (Viewpoint Control Policy).
The policy $\pi_\phi$, parametrized by $\phi,$ operates at the perception low-frequency, $\frac{1}{\tau_{h}}$, and computes informative viewpoints 
{$\va_t$ }for the drone.
The viewpoints are then tracked with an MPC controller that generates the low-level control inputs, $\vu_t$, at the necessary higher frequency, $\frac{1}{\tau_{l}}$.
A positive side-effect of this decomposition in two 
{temporal abstraction levels} is the possibility to neglect the complex drone dynamics in the POMDP formulation used for the RL algorithm.
An overview of the proposed framework is given in Figure \ref{fig:overview}.
To simplify the 
{notation, in this section} $t$ denotes time periods of $\tau_{h}$, 
{whereas the faster time steps of periods $\tau_l$ are denoted by $k$ in Section \ref{sec:MPC}.}

\subsection{Target class observations and Belief Updates} \label{sec:method_belief_updates} 

An important aspect to consider is how to aggregate the different observations made by the drone about each target's class to produce the class beliefs.
The first issue to consider is that standard Bayesian recursive estimation is not advisable because the measurement likelihood model for the update, $\mathbb{P}(\vp_t^j | \vb_{t-1})$, is not available 
from a black-box sensor.
\newcont{To train and learn an accurate pose-dependent model of the likelihood, a dense dataset must be built first. Then, to use it for optimal viewpoint search all targets and their occlusions must be considered. This process is costly and scales badly.}



Instead, in this paper we propose to use a mathematical method called conflation~\cite{Hill2011}. Conflation is used to aggregate probability distributions obtained from measurements over the same phenomena under different circumstances. Notably, this technique has the property of minimizing the loss of Shannon information when flattening multiple independent probability distributions into a single one, that is, computing $\vb_t^j$ given the measurements $\vp_{0:t}^j.$ In addition, it is a commutative and associative operator, enabling easy and efficient recursive computation and making it appealing for onboard computation.
The conflation is defined by
\begin{equation}\textstyle
\label{eq:conflation}
    {\vb}^j_{t} = 
    \zeta(\vp^j_{1:t})
    \equiv \zeta({\vb}^j_{t-1},\vp^j_{t})
    = 
    \frac{\vb_{t-1}^j \odot \, \vp_t^j}{(\vb_{t-1}^j)^{\!\top} \vp_t^j}
    \,,
\end{equation}
where the Hadamard product $\odot$ in the numerator is taken component-wise, whereas the dot product is the normalization factor.
Beliefs are initialized at $t=0$ with a uniform prior over 
{classes in $\mathcal C$}.



Lastly,  although Eq.~\eqref{eq:conflation} considers all the measurements equally, it can easily be extended to a weighted version using the weights as powers of the probability distributions. This could be useful for example in cases where the black-box also outputs a confidence measurement over the class probabilities.

\subsection{Viewpoint Control Policy}\label{sec:policy_network}
\subsubsection{POMDP Formulation}
We formulate the high-level viewpoint recommendation problem as a POMDP, which is defined by the tuple $\langle S, A, \cT, Z, \cO, R \rangle$. 
{The states $\vs_t \in S$ contain the drone's position $\vq_t$ and yaw orientation $\psi_t$, all targets' states $\vy^j_t$ and ground truth class, and the current beliefs $\vb^j_t$ and $l^j_t, 1 \leq j \leq M$.}
The action space, $A$, models the recommended viewpoints, defined by displacements over the drone's position and yaw, $\va_t = (\Delta \vq_t, \Delta \psi_t)$.
Recommended viewpoints are constrained to a neighborhood and orientation from the current pose, which depends on the maximum distance and angle the drone can traverse in $\tau_h$ seconds. 
The transition probability function, $\cT$, simply assumes that the drone is able to reach the output viewpoint in time for the next measurement.
The drone observes partial environment information $\vo_t \in Z$, according to the observation function $\cO$. This observation is defined by $\vo_t = \{\vo_t^j\}_{j=1,...,M}$, where $\vo_t^j \coloneqq (\mathbf{y}^j_t, \mathcal{H}[\vb^j_t], \mathcal{H}[\vp^j_t], l^j_t)$ is the information available from target $j$ regarding its relative pose, velocity, normalized entropy of the belief and the current measurement, and whether it has already been classified. The use of the belief and measurement entropy, instead of the probability distribution vector, enables handling an arbitrary number of classes $C$ and keep track of how much information 
{can still be gained by observing} a target. 
We also recall that for unobserved targets, $\vp_t^j$ is a uniform distribution.


Finally, the reward function, $R$, is shaped based on the problem described in Eq.~\eqref{problemformulation}, and additional factors that the policy should take into account.
First of all, there is a dense reward, $R_{\mathcal{H}}$, proportional to how much the entropy of each belief, $\vb^j_t$, has decreased each time step, $t$, due to the new information gathered. Additionally, there is one sparse reward, denoted by $R_{l}$, for individual target classifications, when any $l_t^j$ changes from zero to one at time $t$, and another, $R_{\mathcal{J}}$, for completing the classification of all of the existing targets, when $\sum_j l_t^j=M$.
On the other hand, to promote solving the task quickly and efficiently, the 
reward includes a constant penalty associated to the time required to classify the targets, $R_{t}$, and another one proportional to the distance to the recommended viewpoint, $R_{a}$, to favor small motions.
The formal definition of all the reward terms is
%

\vspace{-2mm}
\begin{eqnarray*}
    R(\vs_t,\va_t, \vs_{t+1})&\!\!\!=\!\!\!& R_{\mathcal{H}}(\vs_t, \vs_{t+1}) + R_{l}(\vs_t, \vs_{t+1}) \\
    && + R_{\mathcal{J}}(\vs_{t+1}) 
    - R_{t} - R_{a}(\va_t),
\end{eqnarray*}
\vspace{-1mm}
\begin{eqnarray}
        \nonumber
        \text{where}~~ 
        &&R_{\mathcal{H}}(\vs_t, \vs_{t+1}) = w_{\mathcal{H}}\sum^{M}_{j=1}\Big(\mathcal{H}(\vb^j_{t})-\mathcal{H}(\vb^j_{t+1})\Big),\\
        \label{rew:fullrew}
        &&R_{l}(\vs_t, \vs_{t+1}) = w_{l}\sum^{M}_{j=1}(l^j_{t+1}-l^j_{t}),\\
        \nonumber
        &&R_{\mathcal{J}}(\vs_{t+1})= 
      \begin{cases}
      w_{\mathcal{J}} & \hbox{if } \sum_{j=1}^M l^j_{t+1} = M\\
      0 & \hbox{otherwise}\\
     \end{cases},\\
    \nonumber
    &&R_{t} = w_{t},\\[1mm]
    \nonumber
    &&R_{a}(\va_t) = w_{a}(\|\Delta\vq_t\| + |\Delta\psi_t|), 
\end{eqnarray}
\vspace{-2mm}\par\noindent
with $w_{\mathcal{H}}$, $w_{l}$, $w_{\mathcal{J}}$, $w_{t}$ and $w_{a}$ weights that scale each term.

\begin{figure}[t!]
    \centering
    \includegraphics[width=0.49\textwidth]{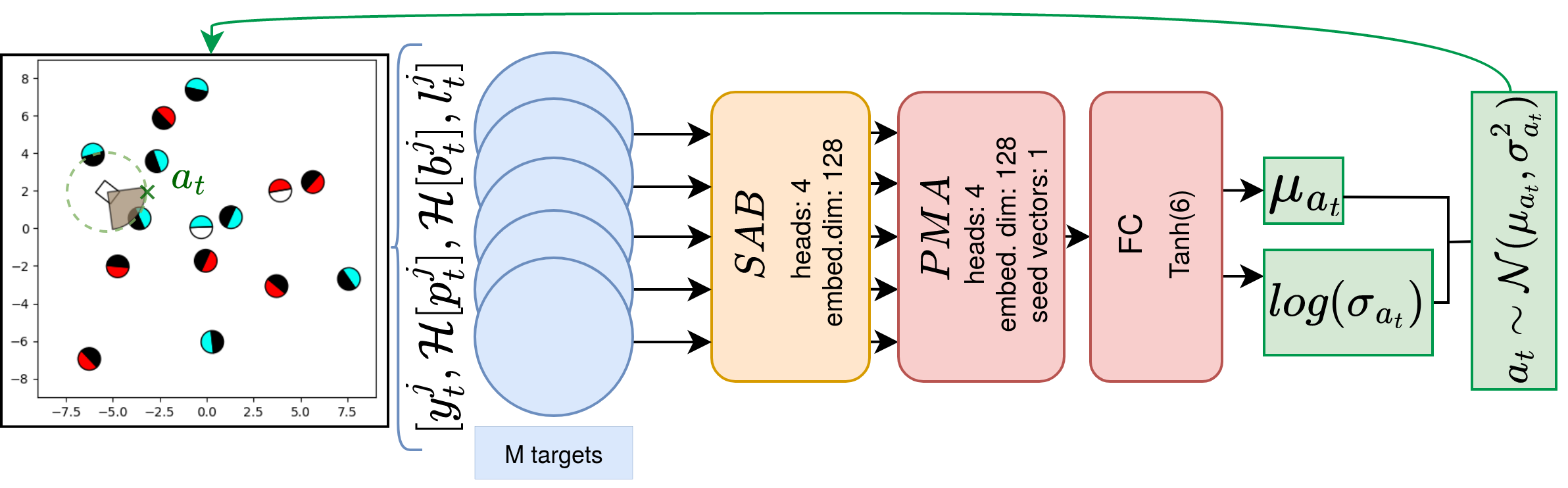}
    \vspace{-4mm}
    \caption{\footnotesize{Policy neural network architecture. The sequence of target information is fed to the self-attention block (SAB) which encodes and identifies how each target pose affects the visibility of the others from the drone's perspective. This information is fed to the pooling multi-head attention layer (PMA) and mapped through a fully connected layer (FC) to obtain the policy viewpoint recommendation.}}
    \label{fig:architecture}
\end{figure}
\vspace{2mm}
\subsubsection{Architecture}
The ability of any learned policy $\pi_\phi(\va_t|\vo_t)$ to generalize beyond the exact situations seen during training, e.g., more targets or changing target behaviors, depends crucially on the chosen neural architecture, shown in Figure~\ref{fig:architecture}. We arrange the information available of each target, $\vo^j_t,$ into a set $\vo_t$. 
The main challenge arises from the set being large and changing over time. 
Inspired by Relational Graph Convolutional Networks~\cite{10.1007/978-3-319-93417-4_38} and self-attention mechanisms~\cite{NIPS2017_3f5ee243} used on static knowledge graphs, we implement a self-attention block (SAB) 
to identify the relationship between the poses of all targets, i.e.,~at time $t$. Thus, the first layer is, 

\begin{equation}
    \begin{aligned}
    & \tilde{\ve}^{1,h}_i = F(\vo_t^i;\mathbf{W}_{q,h}^1)+ \sum\limits_{j\in\mathcal{J}}\lambda^h_{i,j}F(\vo_t^j;\mathbf{W}_{v,h}^1) \,, \\
    & \ve^{1}_i = LN(Res^1(LN(concat(\{\tilde{\ve}^{1,h}_i\}_{h=1...H})))) \,, \\
    & \lambda_{i,j}^h=\text{softmax}\Big(\frac{1}{\sqrt{d_{h}}} F(\vo_t^{i};\mathbf{W}_{q,h}^{1})^\top F(\vo_t;\mathbf{W}_{k,h}^{1}) \Big)_j \,,
    \end{aligned}
\end{equation}
where $i\in\mathcal{J}$,{ $Res^l(x) = x+\sigma(F(x;\mathbf{W}^l))$, with} $\sigma$ {being} a ReLU activation function {and} $F$ a parametric affine transformation. $\mathbf{W}^1\in\mathbb{R}^{d_{enc}\times (d_{h}+1)}$ and $\mathbf{W}_{w,h}^{1}\in\mathbb{R}^{d_{h}H\times (d_{in}+1)}, w \in \{v,q,k\},$ are learnable parameters. $d_{in}$, $d_{h}$, $d_{enc}$ are the dimensionality of the input, each head $h$, and the first layer. 
Note that each head $h$ encodes a different relation $\bm\lambda^h$ between targets.
The purpose of this first layer is to encode information such as target visibility, perspective from which each target's information can be observed, occlusions, and possible simultaneous observations.

Next, we draw inspiration from Set Tranformers~\cite{lee2019set}, used in static set-structured data, to aggregate the features of all latent target representations. We employ a pooling multi-head attention mechanism (PMA) where 
a learned seed vector {per head $\vv_s^h \in \R^{d_h}$} is employed to compute the attention weights for a single query,


\begin{equation}
    \begin{aligned}
    & \tilde{\ve}^{2,h} =\vv_s^h+ \sum\limits_{j\in\mathcal{J}}\lambda^h_{j}F(\ve^1_j;\mathbf{W}_{v,h}^2) \,, \\
    & \ve^{2} = LN(Res^2(LN(concat(\{\tilde{\ve}^{2,h}\}_{h=1...H})))) \,, \\
    & \lambda_{j}^h=\text{softmax}\Big(\Bigl\{\frac{1}{\sqrt{d_{h}}}\vv_s^{h,\top}\wbr{\mathbf{W}_{q,h}^{2,T}}{}F(\ve^1_j;\mathbf{W}_{k,h}^{2})\Bigl\}_{j\in\mathcal{J}} \Big)_j \,.
    \end{aligned}
\end{equation}

This results in a latent vector $\ve^2$ that is mapped, through a fully connected layer, to the parameters $\mu_{\va_t}$ and $\log(\sigma_{\va_t})$ of a diagonal Gaussian distribution $\mathcal N(\mu_{\va_t}, \sigma_{\va_t})$ over viewpoints. The learned policy $\pi_\phi$ samples recommended viewpoints $\va_t$ from this distribution.
We use the Proximal Policy Optimization (PPO) algorithm to train the network~\cite{schulmanPPO,liang2018rllib}. As learning algorithms like PPO also require an estimate of the state-value $V^{\pi_\phi}\wbr{_\chi}{}(\vs_t) = \mathbb{E}[\sum^\infty_{t'=t} \gamma^{t'-t} R(\vs_{t'},\va_{t'})|\vs_{t'}\sim\cT,\va_{t'}\sim\pi_\phi]$, where $\gamma$ is the discount factor, another linear layer predicts $V^{\pi_\phi}(\vs_t)\approx \vv_v^\top \ve^2$. The latter is only used during training to guide the policy. We combine both the surrogate loss and KL-divergence term to stabilize training. We also use an entropy regularization term to encourage exploration \cite{Haarnoja17}. We refer the reader to~\cite{schulmanPPO} for more information on the algorithm equations and details. 






\subsection{Low-level Controller} \label{sec:MPC}
To account for the drone dynamics, the recommended viewpoint position $\va_t$ is tracked with an MPC low-level controller~\cite{Zhu2019}. 
Let $\vx_{at}$ denote the viewpoint state output by the policy in the world frame, obtained from the current drone state and setting to zero the information that is not considered in $\va_{t},$ e.g., roll and pitch.  
Every $\tau_l$ seconds, the controller solves the following receding horizon constrained optimization problem,
\begin{equation} \label{lowlevel_control}
    \begin{aligned}       \min_{\vx_{1:N},\vu_{0:N-1}} & \sum_{k=0}^{N-1} J^k(\vx_k, \vu_k) + J^N(\vx_N, \vx_{at})\\
        \text{s.t.} \quad & \vx_0 = \vx_t,  \quad \vx_{k+1}=f(\vx_k,\vu_k)\\
        & \vu_{k} \in \cU, \quad 0 \leq k \leq N\!-\!1
    \end{aligned}
\end{equation}
where $\vu_k$ is the low-level control input sent to the robot, that needs to be inside the possible values $\cU$, $f(\vx^k,\vu^k)$ the internal dynamics and $J^k(\vx_k,\vu_k) = w_{u}\norm{\vu_k},$
\vspace{1mm}
\begin{equation}
    J^N(\vx_N, \vx_{at}) = w_g\frac{\norm{\vx_N-\vx_{at}}}{\norm{\vx^0-\vx_{at}}},
\end{equation}
\vspace{-2mm}\par\noindent
the stage and terminal costs, weighted by $w_{u}$ and $w_{g}$ respectively. For more details we refer the reader to~\cite{Zhu2019}. 

\section{Implementation Details}\label{sec:implementation}
We train and test the proposed method both with simulated perception  
and with a real classifier in a photo-realistic simulator. The first environment has a simplified, computationally efficient observation model and is used for comparison between our method and the baselines introduced in Section \ref{sec:baselines}. 
The second environment is used to test the proposed method under more realistic conditions.

\subsection{Simulated Perception Environment} \label{sec:toyEnv}

\subsubsection{Observation Model}
We consider a synthetic pinhole camera with focal length equal to 400 pixels that acquires images of $640\times 480$ pixels. A target is modeled as visible if we can draw a line between its center and the drone's without any collision.
For each target we consider only the half part of the cylinder that is facing forward and project it into the image frame, if it is visible from the camera perspective.
This generates an image of a trapezoid with area and skew depending on the relative position and orientation of the target
~(Figure \ref{fig:overview}).
We use these two parameters to determine the probability distribution of the observation over the class set, decreasing the probability of the true class exponentially with the skew and increasing it linearly with the area.
We also penalize heavily trapezoids that do not fit in the image.
Our method does not have any knowledge of neither the observation model nor the classification model, which makes them black-box to it without loss of generality. 

\subsubsection{Training conditions} \label{sec:trainingsetup_and_env}
Our high-level policy and the learned baselines are trained in closed environments of $50\times50~m^2$ with simulation steps of $\tau_h = 0.25~s$. As shown in Figure \ref{fig:overview}, targets are modeled as cylinders of $0.6~m$ radius, $1.8~m$ of height, with their class information only visible from the front. The targets follow constant velocity dynamics and belong to either class \textit{red} or class \textit{blue}. Targets' speed in every axis is sampled around 1~m/s and clipped at 1.5~m/s. Whenever targets collide among themselves or against walls, they rebound conserving kinetic energy and momentum.

The drone's maximum angular and linear speed in each axis is respectively $\dot\psi^{max} = 60^o/s$ and 2~m/s. This is to ensure that the drone can reach targets moving away from it. 
To reduce computation costs, only during training we assume that the drone follows first-order dynamics, and control its velocity to guide the drone to the recommended viewpoint. 

Each episode, each method is given $100$ seconds to classify all targets ($l^j_t=1$, $\forall j \in \cJ$, $b_{\max}=0.95$). 
Episodes are finished after reaching the timeout, or successfully classifying all targets. Values for the reward weights are $w_{\cJ}=100$, $w_{l}=5$, $w_{\cH}=1$, $w_{t}=0.3$, $w_{a}=0.01$.

\subsection{Training Algorithm}

\begin{table}[t]
    \centering
    \caption{\footnotesize{Hyperparameters for PPO training algorithm}}
    \color{black}\begin{tabular}{|c|c|}
    \hline
        Parameter & Value \\
    \hline
        Time steps each update & 16000 \\
        SGD minibatch size & 256 \\
        Learning rate & 3e-4 \\
        Entropy loss coeff. $c_2$ & 0.001 \\
        Gradient Clipping & 0.1 \\
    \hline
    \end{tabular}
    \label{table:hyperparam}
\end{table}

\begin{figure*}[t!]
        \captionsetup[subfigure]{position=b}
        \begin{subfigure}{0.32\textwidth}
                \includegraphics[width=\textwidth]{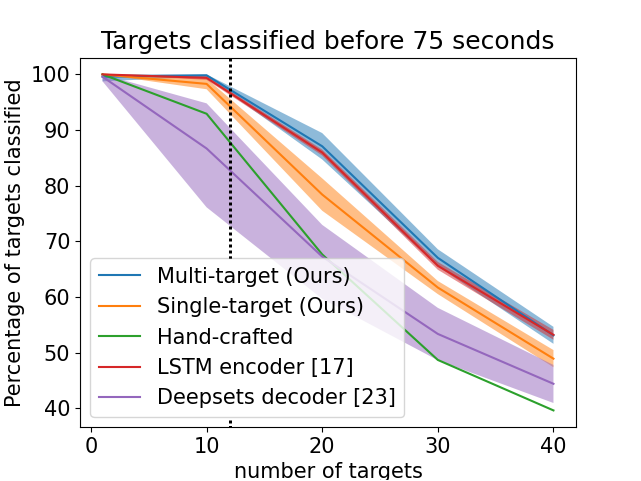}
        \vspace{-1mm}
        \end{subfigure}
        ~
        \captionsetup[subfigure]{position=b}
        \begin{subfigure}{0.32\textwidth}
                \includegraphics[width=\textwidth]{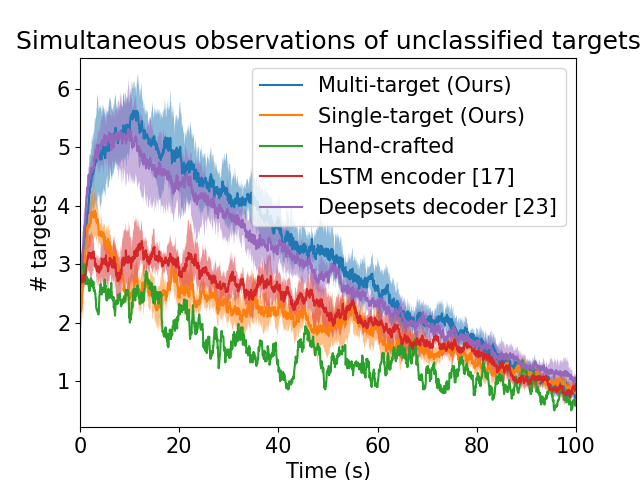}
        \vspace{-1mm}
        \end{subfigure}
        ~
        \captionsetup[subfigure]{position=b}
        \begin{subfigure}{0.32\textwidth}
                \includegraphics[width=\textwidth]{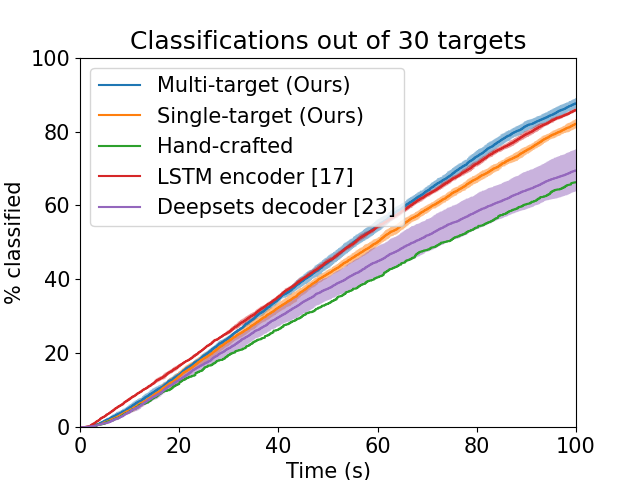}
        \vspace{-1mm}
        \end{subfigure}
        ~
        \captionsetup[subfigure]{position=b}
        \begin{subfigure}{0.32\textwidth}
                \includegraphics[width=\textwidth]{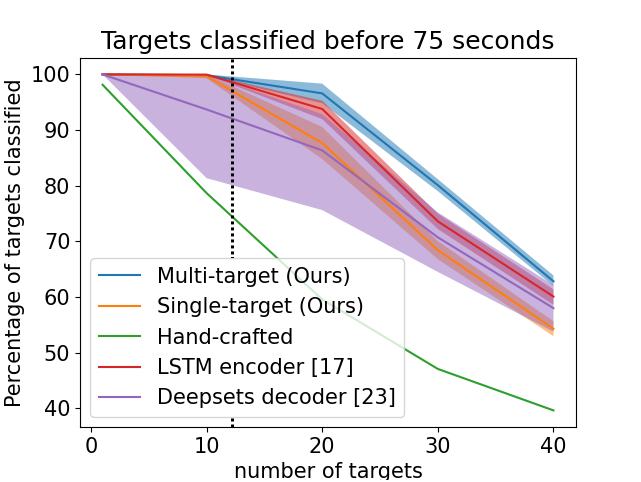}
        \vspace{-1mm}
        \end{subfigure}
        ~
        \captionsetup[subfigure]{position=b}
        \begin{subfigure}{0.32\textwidth}
                \includegraphics[width=\textwidth]{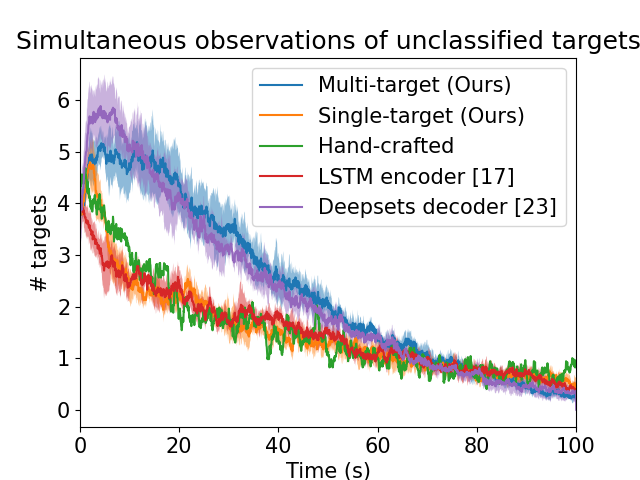}
        \vspace{-1mm}
        \end{subfigure}
        ~
        \captionsetup[subfigure]{position=b}
        \begin{subfigure}{0.32\textwidth}
                \includegraphics[width=\textwidth]{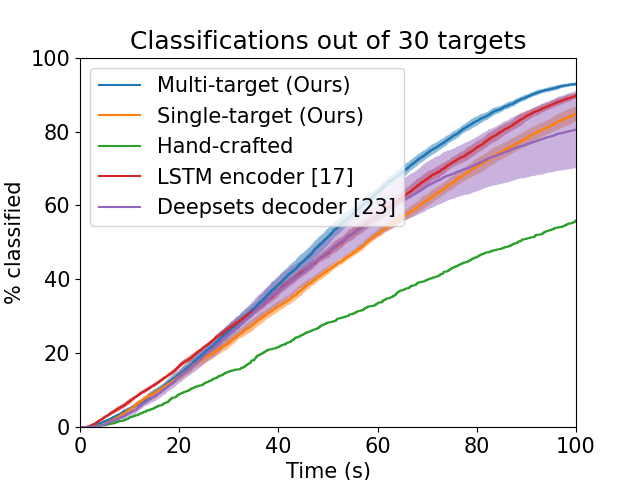}
        \vspace{-2mm}
        \end{subfigure}
        
        \vspace{-3mm}
        \caption{\footnotesize{\textbf{Top row)}~Experiments with targets following constant velocity dynamics, as in training.~\textbf{Bottom row)}~Experiments with targets following social forces dynamics\newcont{ not seen during training}.~\textbf{Left)}~Comparison of the percentage of targets classified before timeout in environments with 1 to 40 targets.\newcont{ The black line denotes the maximum amount of targets seen during training.}~\textbf{Center)}~Evolution of simultaneous observations along the episode in environments of 30 targets.~\textbf{Right)}~Classification speed in environments of 30 targets.
        }} \label{fig:firstperformance}
        \vspace{-8mm}
\end{figure*}


\rebuttalsub{The hyperparameters of the PPO algorithm used for training are shown in Table \ref{table:hyperparam}\newcont{. We refer the reader to~\cite{schulmanPPO,Engstrom2020Implementation,andrychowicz2021what} for a detailed analysis on the effect of these hyperparameters on the training algorithm and how to tune them.} The learning algorithm and the training of our policy are implemented
, using the RLlib framework~\cite{liang2018rllib}.
 }
 {
 The learning algorithm and the training of our policy are implemented using the RLlib framework~\cite{liang2018rllib}. Table~\ref{table:hyperparam} shows the hyperparameters tuned for training the PPO algorithm. We tuned these hyperparameters because they regulate the speed and quality of training and are more problem-specific. Other hyperparameters follow the default values provided by the framework. We refer the reader to~\cite{schulmanPPO,Engstrom2020Implementation,andrychowicz2021what} for a detailed analysis on the effect of these hyperparameters on the training algorithm and how to tune them.
 }
 State space exploration is enhanced
 ~\cite{domainRandomization} to make the learned policy more robust at test time, by randomizing the drone's initial poses, and the targets' initial poses, velocities and beliefs over their classes. Some of the targets are randomly selected and enforced to remain static.

We follow a two-phase training schedule. First, we train the policy for the general case in scenarios, chosen randomly, containing between 1 and 12 targets. Then we fine tune the resulting policy by training it in scenarios containing only 1 to 6 targets. While counter-intuitive, this choice of schedule is motivated by how the task of active classification changes depending on the number of targets present in the environment. There are two motion planning tasks that need to be solved to perform active classification: simultaneous observations of multiple targets and, if that is not possible, then focus on single target high-quality viewpoints. We 
{find} that only learning both tasks simultaneously in environments with 1-12 targets results in a policy prioritizing simultaneous observations of multiple targets. Such policy scales up well but at the expense of poor choice of viewpoints when the number of targets is small. This is why a second training phase to fine-tune the policy to environments requiring single-target viewpoints is used.


The policy and value function (detailed in section \ref{sec:policy_network}) were trained for $9.6e7$ steps in environments containing up to twelve targets, sampled uniformly (first phase). Then, they were further trained for $3.2e7$ steps in environments containing a random number between one to six targets (second phase). The training of the algorithm was done using an Intel i9-9900 CPU@3.10GHz computer. 

Computing the policy actions takes less than $4ms$ per time step, which allows for a real-time implementation of the framework with a fast control frequency.

\section{Simulated Results} \label{sec:results}
We analyze our high-level policy's scalability and robustness to different target dynamics in comparison with other hard-coded and learned baselines.




\subsection{Baselines}\label{sec:baselines}
There are no existing approaches that address the problem of active classification of dynamic targets without relying on an available observation model. Therefore, the baselines used for comparison are:
\begin{itemize}        
    \item \textbf{Hand-crafted}: Hard-coded policy that guides the drone to a position \qty{2}{\meter} in front of an unobserved target, pointing the camera at it.
    \item \textbf{Single-target (Ours)}: This is an ablation of our method. A learned single-target policy is given the information of an unobserved target and guides the drone to viewpoints allowing it to classify it as fast as possible. It is trained in scenarios with just one target. 
    \item \textbf{LSTM encoder~\cite{meverett2021,britogompc}}: The self-attention and attention pooling layers are substituted by a linear layer followed by an LSTM. Every unclassified target is inversely sorted by proximity to the drone. The first layer encodes each target's information, and the sequence is fed into the LSTM.
    \item \textbf{DeepSets decoder~\cite{Hsu2021}}: We replace the attention pooling layer by a mean pooling layer. 
\end{itemize}
\newcont{The latter two baselines are recently proposed scalable architectures employed in different, albeit similar~\cite{Hsu2021}, problems.}

\subsection{Test conditions}
At test time, the whole pipeline described in section \ref{sec:method} is used, including the drone dynamics and controller described in section \ref{sec:MPC}. 
We evaluate our method's robustness under target behavior both seen and unseen during training. Aside from constant velocity dynamics, we test our policy in environments where targets 
follow social forces pedestrian dynamics~\cite{socialforces}. As in training, each method is given up to 100 seconds to classify all targets. Each policy is trained with 5 different seeds. The results of all seeds are averaged and shown with their standard deviations. For every test, we evaluate and average each method's results over 50 episodes. 
Initial conditions of all evaluation episodes are randomised but maintained equal across our method and all baselines.


Since the first two methods are designed to classify one target
, the extension to multiple targets is done through their sequential classification. Unclassified targets are ordered by their distance to the drone. The drone classifies the first in the list before moving on to the next one. Observations of other targets are still accounted for in the information fusion and belief computation.


\subsection{Scalability analysis} \label{sec:results:scalability}
We evaluate each method in environments containing up to 40 targets to test their scalability to larger number of targets than seen during training. In Figure~\ref{fig:firstperformance}\textbf{-Left} , we report each method's percentage of classified targets at the 75 second mark in environments containing different numbers of targets. 
In general, as expected, there is a drop in the percentage of targets that can be classified 
when their quantity increases. This is due to the task theoretically requiring more steps 
and an increase of occlusions generated by other targets. The results show that our method is the one able to generalize best to target dynamics unseen during training and generally outperforms all other baselines in environments containing much larger amounts of targets than seen during training.\newcont{ While our method does not take any assumption on how target information should be weighted, \cite{britogompc} relies heavily on the priority order given to targets, as analysed in~\cite{meverett2021}. This is the reason why, in our experiments,~\cite{britogompc} shows similar scalability results under target dynamics seen during training (Figure~\ref{fig:firstperformance}, \textbf{Top row}) but does not generalize as well as our method to unseen dynamics (Figure~\ref{fig:firstperformance}, \textbf{Bottom row}).}




\subsection{Policy behavior}
We provide an empirical analysis on each method's behavior and the effect on its performance. In Figures~\ref{fig:firstperformance}\textbf{-Center} and~\ref{fig:firstperformance}\textbf{-Right}, 
we test our policy in 
{simulated perception }environments with 30 targets and report the number of unclassified targets simultaneously observed along the episode. During the first half of the episode, our method consistently is shown to observe and provide classification estimates of more targets than other baselines, which results in faster classification of targets. This shows that our method is able to learn the effect of each target on the observations of others and discover groups of simultaneously observable targets. The latter is difficult to achieve by single-target classification baselines or methods that assume distance-based relations among targets. Similarly, 
{the lower performance albeit similarly high simultaneous observations} of the \textit{DeepSets decoder} baseline, especially at the end of the episode when there are less targets to classify, suggest that the attention-based pooling layer allows better aggregation of each target's latent information, effectively establishing a classification priority order. 
The sharp decline in simultaneous observations of unclassified targets is due to simultaneously observable targets becoming classified, which results in a sparser distribution of unclassified targets across the environment. 

\section{Photo-Realistic Results} \label{ref:relResults}
\setcounter{figure}{5}
\begin{figure*}[b!]
        \vspace{-4mm}
        \hfill
        \captionsetup[subfigure]{position=b}
        \begin{subfigure}{0.30\textwidth}
                \includegraphics[width=\textwidth]{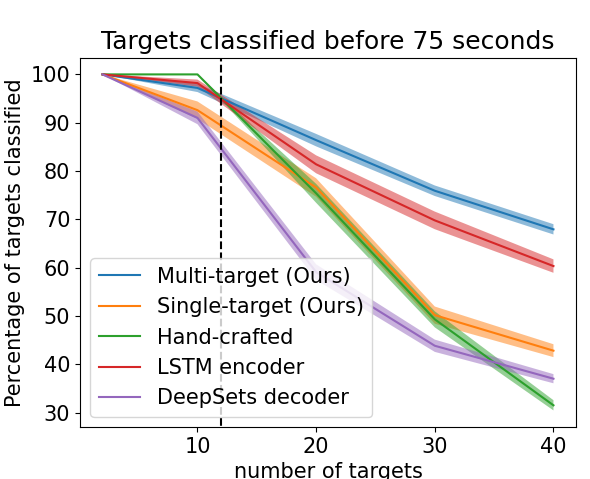}
        \vspace{-5mm}
        \caption{\footnotesize{}\label{fig:scalabilityReal}}
        \end{subfigure}
        ~
        \captionsetup[subfigure]{position=b}
        \begin{subfigure}{0.30\textwidth}
                \includegraphics[width=\textwidth]{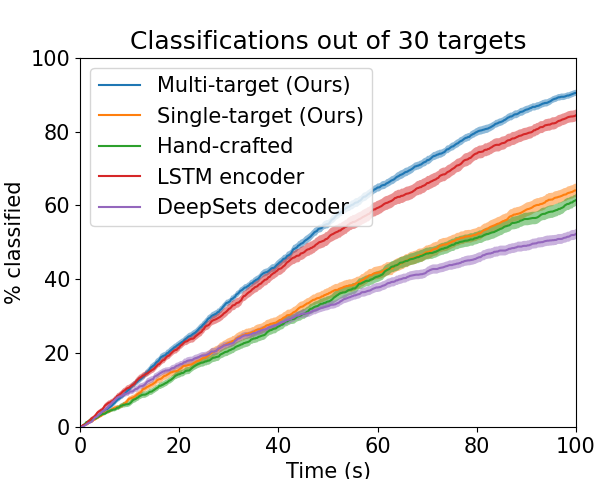}
        \vspace{-5mm}
        \caption{\footnotesize{}\label{fig:classificationspeedRealPerception}}
        \end{subfigure}
        ~
        \captionsetup[subfigure]{position=b}
        \begin{subfigure}{0.30\textwidth}
                \includegraphics[width=\textwidth]{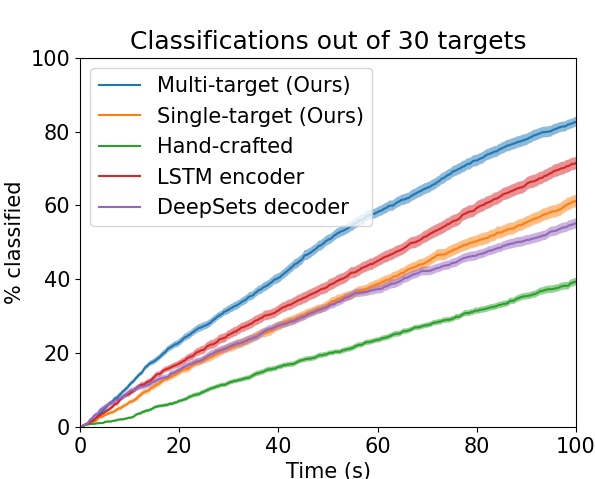}
        \vspace{-5mm}
        \caption{\footnotesize{}\label{fig:classificationspeedReal}}
        \end{subfigure}
        \hfill
        \vspace{-5mm}
        \caption{\footnotesize{\newcont{Mean and standard {error} over 50 experiments in the photo-realistic simulator.
         \textbf{a)}~Comparison of the percentage of targets classified before timeout of 75 seconds in environments with 1 to 40 targets, where measurements are directly taken from viewpoints that the DRL policy suggests. \textbf{b)} Classification speed in the experiment of (a) with 30 targets.
          ~\textbf{c)}~Classification speed in photo-realistic environments of 30 targets with realistic drone dynamics and AirSim's controller.
        }}} \label{fig:relperformance}
\end{figure*}
\setcounter{figure}{3}

\begin{figure}[t]
    \centering
    \includegraphics[width=0.48\textwidth]{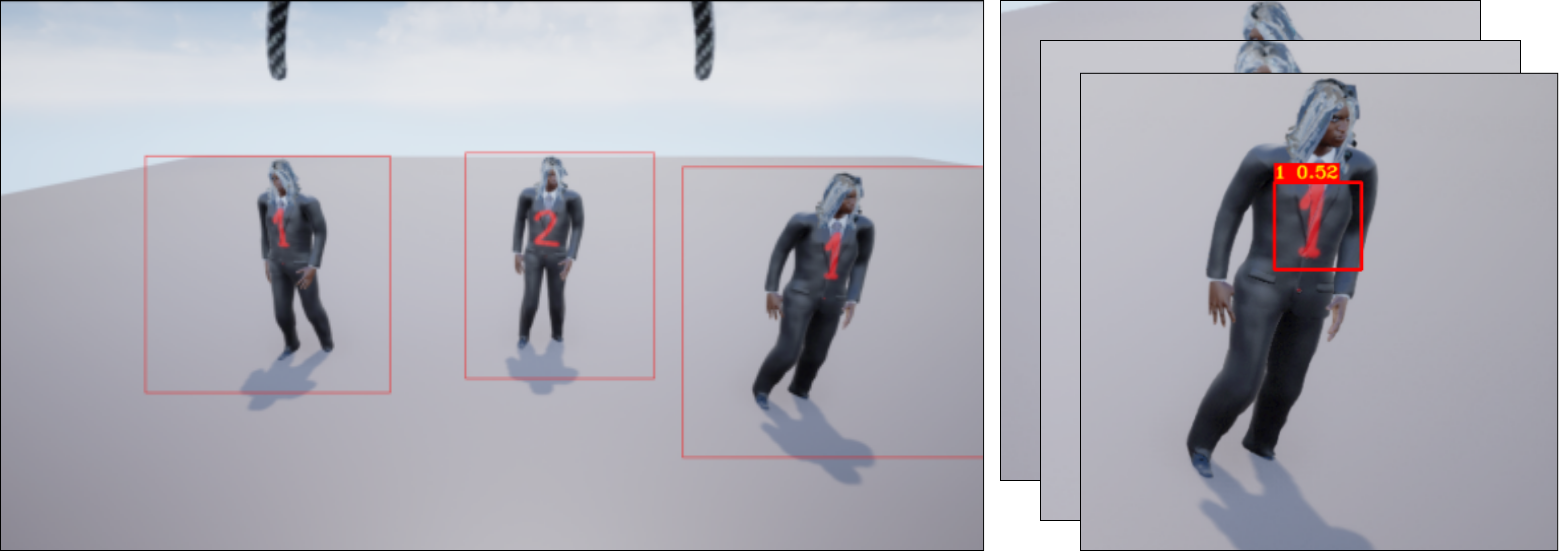}
    \caption{\footnotesize{Pedestrian models, as seen from the drone's perspective in the realistic environment. Each pedestrian is tracked using AirSim's API. 
    Its class is represented by a red number in the front, classified using YoloV3.
    }}
    \label{fig:Uclassped}
    \vspace{-1mm}
\end{figure}

To ascertain our framework's capabilities under realistic conditions, 
we test its performance in an environment generated with Unreal Engine using AirSim~\cite{airsim} to simulate drone control and perception. We simulate our targets using open-source 3D human meshes, produced in \textit{MakeHuman} \cite{makehuman}, which move to random goals while avoiding collisions in an environment of \rebuttalsub{$24.9\times21.2~m^2$}{$50\times50\,m^2$}. As shown in Figure \ref{fig:Uclassped}, to remain close to the use-case shown in the earlier simpler environment, pedestrian classes are represented by a red number in the front of their shirt. 

\subsubsection{Photo-Realistic Observation Model}
We obtain the cropped image of every pedestrian detected in the drone's field of view (FOV) using the AirSim API, avoiding the problem of data association. Each image is resized, and everything other than the painted number is filtered out.

An implementation of the YoloV3~\cite{yolov3} algorithm is used to detect and classify the digit in each processed image. We train the algorithm using a dataset of rotated, up-scaled and down-scaled MNIST images. Both YoloV3 implementation and the adapted dataset have been obtained from \cite{pythonlessons}.

The output of the algorithm consists in a set of bounding boxes, each one associated to a detected digit and a normalized score stating how sure the algorithm is of it's detection. We rule out those boxes containing a number of pixels smaller than a threshold $B_a = 2000$ and take the box with the highest score. 
The normalized score of the detected digit is used as the target class probability estimate, distributing the remaining probability mass among the other classes. The output is a uniform distribution when no digits are detected, or the normalized score is below the uniform distribution probability.

\subsubsection{Training conditions} \label{sec:relEnvTraining}
We use the training setup explained in Section \ref{sec:toyEnv}, using the realistic observation model.
Nevertheless, to avoid the computational cost of running Unreal/Airsim and YoloV3, for training we use AirSim to extract a dense library of pedestrian observations of all classes in different relative poses from the drone's FOV. For each class, each pose-dependent image is used to compute and save a probability distribution. During training, for each target in the drone's FOV, we use the probability distribution of the library image whose pose is the closest to the target's current relative pose.

\begin{figure}[t] 
        \captionsetup[subfigure]{position=b}
        \begin{subfigure}{0.10\textwidth}
                \includegraphics[width=\textwidth]{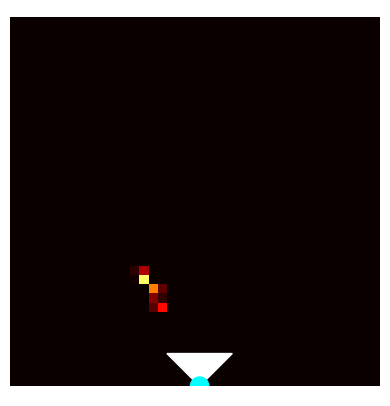}
        \caption{\footnotesize{-100$^o$}\label{fig:-100}}
        \end{subfigure}
        ~
        \captionsetup[subfigure]{position=b}
        \begin{subfigure}{0.10\textwidth}
                \includegraphics[width=\textwidth]{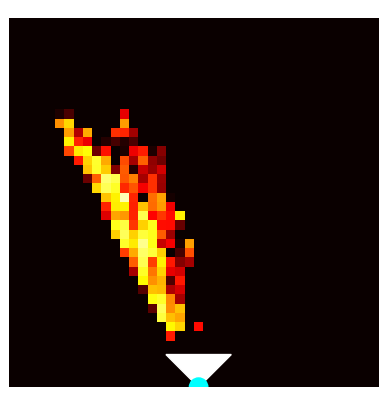}
        \caption{\footnotesize{-140$^o$}\label{fig:-140}}
        \end{subfigure}
        ~
        \captionsetup[subfigure]{position=b}
        \begin{subfigure}{0.10\textwidth}
                \includegraphics[width=\textwidth]{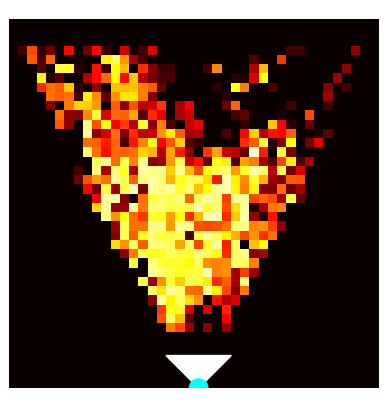}
        \caption{\footnotesize{$\pm$180$^o$}\label{fig:-180}}
        \end{subfigure}
        ~
        \captionsetup[subfigure]{position=b}
        \begin{subfigure}{0.13\textwidth}
                \includegraphics[width=\textwidth]{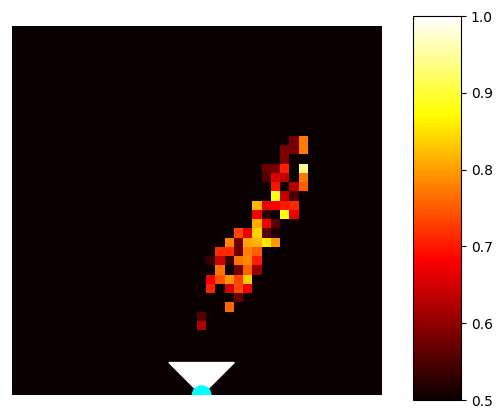}
        \caption{\footnotesize{140$^o$}\label{fig:140}}
        \end{subfigure}
        \vspace{-8.5mm}
        \caption{\footnotesize{Probability heat-map of the employed black-box classifier for a target of class 1. For a robot placed at the bottom, facing upwards and a FOV of $90^o$, every plot shows the output probability of identifying the real class of the target at different relative positions and a fixed relative orientation from the robot.
        }} \label{fig:ObsModel}
        \vspace{-1mm}
\end{figure}

Being in a controlled environment, during training we monitor the output of the perception system to 
detect classification errors, and substitute them by a uniform probability distribution. 
This choice enables our policy to prioritise good over bad classification viewpoints, adding an additional robustness mechanism to outlier classifications to the control policy.
An example of the resulting normalized classification score is given in Figure \ref{fig:ObsModel}.

\subsubsection{Real Perception Environment} \label{sec:relEnv}
This time we test our method's performance in the photo-realistic environment. 
In Figures \ref{fig:scalabilityReal} and \ref{fig:classificationspeedRealPerception}, we only analyze the performance of the viewpoint recommendation policy without having AirSim's different drone dynamics and in-built controller affecting the results, which we use for comparison in Figure \ref{fig:classificationspeedReal}. We show results for our method and how it compares \rebuttalsub{to the \textit{Hand-crafted} policy and the \textit{Single-target} ablation in this new setting}{in this new setting to the baselines presented in Section~\ref{sec:baselines}}.

As in section \ref{sec:results}, we evaluate each method 50 times in environments containing up to 40~targets for 75 and 100~seconds, and show the resulting mean and standard {error}. In all our experiments, the mean percentage of misclassifications was under $3\%$ with no significant differences among different methods. Figure \ref{fig:relperformance} shows that the results presented in Section \ref{sec:results} hold in the new photo-realistic environment, even when realistic drone dynamics and control are used. \rebuttalsub{Note that the higher performance in comparison to Figure \ref{fig:firstperformance} is due to the smaller environment and the fact that we take measurements directly from the recommended viewpoints.}{Note that the different performance in comparison to Figure \ref{fig:firstperformance} is due to the different observation model and the fact that we take measurements directly from the viewpoints recommended by the DRL policy in Figures \ref{fig:scalabilityReal} and \ref{fig:classificationspeedRealPerception}.} This is very relevant from the Sim-to-Real perspective, considering that the policy has been trained in a different environment with a perception system approximated from the one used during testing.







\section{Conclusion}
In this paper, we have introduced a framework for active classification of multiple dynamic targets when the information is extracted using a ``black-box'' classifier. The proposed framework learns a policy that outputs viewpoints through Deep Reinforcement Learning using an attention-based, permutation invariant architecture. 
Then, a low-level MPC controller moves the drone to the viewpoints taking care of the complex dynamics at high-frequency.
Sensor fusion of the black-box sensor is done through conflation.
The results have shown that our policy outperforms multiple baselines, \rebuttalsub{is able to generalize to scenarios not seen during training, and scales to environments with more than double the amount of targets experienced during training.}{both in terms of generalization to target dynamics not seen during training and scalability to environments with more than double the amount of targets experienced during training.}
However, there is a limit to the number of targets one robot can classify under given time constraints. In the future, multiple drones could be employed for efficient dynamic active classification.


\appendices

\bibliographystyle{IEEEtran}
\balance
\bibliography{RSS_bib}

\begin{thebibliography}{10}
\providecommand{\url}[1]{#1}
\csname url@rmstyle\endcsname
\providecommand{\newblock}{\relax}
\providecommand{\bibinfo}[2]{#2}
\providecommand\BIBentrySTDinterwordspacing{\spaceskip=0pt\relax}
\providecommand\BIBentryALTinterwordstretchfactor{4}
\providecommand\BIBentryALTinterwordspacing{\spaceskip=\fontdimen2\font plus
\BIBentryALTinterwordstretchfactor\fontdimen3\font minus
  \fontdimen4\font\relax}
\providecommand\BIBforeignlanguage[2]{{%
\expandafter\ifx\csname l@#1\endcsname\relax
\typeout{** WARNING: IEEEtran.bst: No hyphenation pattern has been}%
\typeout{** loaded for the language `#1'. Using the pattern for}%
\typeout{** the default language instead.}%
\else
\language=\csname l@#1\endcsname
\fi
#2}}

\bibitem{Redmon2016YouOL}
J.~Redmon, S.~K. Divvala, R.~B. Girshick, and A.~Farhadi, ``You only look once:
  Unified, real-time object detection,'' \emph{IEEE Conference on Computer
  Vision and Pattern Recognition (CVPR)}, pp. 779--788, 2016.

\bibitem{Alonso-21}
I.~Alonso, L.~Riazuelo, L.~Montesano, and A.~Murillo, ``Domain adaptation in
  lidar semantic segmentation by aligning class distributions,'' in \emph{Int.
  Conf. on Informatics in Control, Autom. and Robot.}, 2021.

\bibitem{Patten2016ViewpointEF}
T.~Patten, M.~Zillich, R.~C. Fitch, M.~Vincze, and S.~Sukkarieh, ``Viewpoint
  evaluation for online 3-d active object classification,'' \emph{IEEE Robotics
  and Automation Letters}, vol.~1, no.~1, pp. 73--81, 2016.

\bibitem{activeWeedClassification}
M.~Popović, G.~Hitz, J.~Nieto, I.~Sa, R.~Siegwart, and E.~Galceran, ``Online
  informative path planning for active classification using uavs,'' in
  \emph{IEEE Int. Conf. on Robotics and Automation}, 2017, pp. 5753--5758.

\bibitem{Natansov2014}
N.~Atanasov, B.~Sankaran, J.~Le~Ny, G.~J. Pappas, and K.~Daniilidis,
  ``Nonmyopic view planning for active object classification and pose
  estimation,'' \emph{IEEE Trans. on Rob.}, vol.~30, no.~5, pp. 1078--1090,
  2014.

\bibitem{Patten2018}
T.~Patten, W.~Martens, and R.~Fitch, ``Monte carlo planning for active object
  classification,'' \emph{Auton. Rob.}, vol.~42, no.~02, pp. 391--421, 2018.

\bibitem{active6D}
J.~Sock, G.~Garcia-Hernando, and T.-K. Kim, ``Active 6d multi-object pose
  estimation in cluttered scenarios with deep reinforcement learning,'' in
  \emph{IEEE/RSJ Int. Conf. on Intelligent Robots and Systems}, 2020, pp.
  10\,564--10\,571.

\bibitem{efficientMultiview}
Q.~Xu \emph{et~al.}, ``Towards efficient multiview object detection with
  adaptive action prediction,'' in \emph{IEEE Int. Conf. on Robotics and
  Automation}, 2021, pp. 13\,423--13\,429.

\bibitem{activePerceptionAutonomousObservation}
D.~Kent and S.~Chernova, ``Human-centric active perception for autonomous
  observation,'' in \emph{IEEE Int. Conf. on Robotics and Automation}, 2020,
  pp. 1785--1791.

\bibitem{ALCANTARA2021103778}
A.~Alcántara, J.~Capitán, R.~Cunha, and A.~Ollero, ``Optimal trajectory
  planning for cinematography with multiple unmanned aerial vehicles,''
  \emph{Robotics and Autonomous Systems}, vol. 140, p. 103778, 2021.

\bibitem{detectionAware}
B.~F. Jeon, D.~Shim, and H.~Jin~Kim, ``Detection-aware trajectory generation
  for a drone cinematographer,'' in \emph{IEEE/RSJ Int. Conf. on Intelligent
  Robots and Systems}, 2020, pp. 1450--1457.

\bibitem{GeometricDL}
M.~M. Bronstein, J.~Bruna, T.~Cohen, and P.~Velickovic, ``Geometric deep
  learning: Grids, groups, graphs, geodesics, and gauges,'' \emph{CoRR}, vol.
  abs/2104.13478, 2021.

\bibitem{DenseCAvoid}
A.~J. Sathyamoorthy, J.~Liang, U.~Patel, T.~Guan, R.~Chandra, and D.~Manocha,
  ``Densecavoid: Real-time navigation in dense crowds using anticipatory
  behaviors,'' in \emph{IEEE Int. Conf. on Robotics and Automation}, 2020, pp.
  11\,345--11\,352.

\bibitem{cui2021}
Y.~Cui, H.~Zhang, Y.~Wang, and R.~Xiong, ``Learning world transition model for
  socially aware robot navigation,'' in \emph{IEEE Int. Conf. on Robotics and
  Automation}, 2021, pp. 9262--9268.

\bibitem{Li2019}
Q.~Li, F.~Gama, A.~Ribeiro, and A.~Prorok, ``Graph neural networks for
  decentralized multi-robot path planning,'' in \emph{IEEE/RSJ Int. Conf. on
  Intelligent Robots and Systems}, 2020, pp. 11\,785--11\,792.

\bibitem{Chenyuying}
Y.~Chen, C.~Liu, B.~E. Shi, and M.~Liu, ``Robot navigation in crowds by graph
  convolutional networks with attention learned from human gaze,'' \emph{IEEE
  Rob. and Autom. Let.}, vol.~5, no.~2, pp. 2754--2761, 2020.

\bibitem{britogompc}
B.~Brito, M.~Everett, J.~How, and J.~Alonso-Mora, ``Where to go next: Learning
  a subgoal recommendation policy for navigation among pedestrians,''
  \emph{IEEE Robotics and Automation Letters}, vol.~6, no.~3, pp. 4616--4623,
  2021.

\bibitem{meverett2021}
M.~Everett, Y.~F. Chen, and J.~How, ``Collision avoidance in pedestrian-rich
  environments with deep reinforcement learning,'' \emph{IEEE Access}, vol.~9,
  pp. 10\,357--10\,377, 2021.

\bibitem{kurin2021my}
V.~Kurin, M.~Igl, T.~Rockt{\"a}schel, W.~Boehmer, and S.~Whiteson, ``My body is
  a cage: the role of morphology in graph-based incompatible control,'' in
  \emph{Int. Conf. on Learning Representations}, 2021.

\bibitem{NIPS2017_f22e4747}
M.~Zaheer, S.~Kottur, S.~Ravanbakhsh, B.~Poczos, R.~R. Salakhutdinov, and A.~J.
  Smola, ``Deep sets,'' in \emph{Advances in Neural Information Processing
  Systems}, vol.~30.\hskip 1em plus 0.5em minus 0.4em\relax Curran Associates,
  Inc., 2017.

\bibitem{Chen2019}
C.~Chen, Y.~Liu, S.~Kreiss, and A.~Alahi, ``Crowd-robot interaction:
  Crowd-aware robot navigation with attention-based deep reinforcement
  learning,'' in \emph{IEEE Int. Conf. on Rob. and Automation}, 2019, pp.
  6015--6022.

\bibitem{NIPS2017_3f5ee243}
A.~Vaswani \emph{et~al.}, ``Attention is all you need,'' in \emph{Adv. in Neur.
  Inform. Processing Systems}, vol.~30, 2017, pp. 1--11.

\bibitem{Hsu2021}
C.~D. Hsu, H.~Jeong, G.~J. Pappas, and P.~Chaudhari, ``Scalable reinforcement
  learning policies for multi-agent control,'' in \emph{IEEE/RSJ Int. Conf. on
  Intelligent Robots and Systems}, 2021, pp. 4785--4791.

\bibitem{lee2019set}
J.~Lee, Y.~Lee, J.~Kim, A.~Kosiorek, S.~Choi, and Y.~W. Teh, ``Set transformer:
  A framework for attention-based permutation-invariant neural networks,'' in
  \emph{Int. Conf. on Mach. Learn.}, 2019, pp. 3744--3753.

\bibitem{sung2017}
Y.~Sung and P.~Tokekar, ``Algorithm for searching and tracking an unknown and
  varying number of mobile targets using a limited fov sensor,'' in \emph{IEEE
  Int. Conf. on Rob. and Autom.}, 2017, pp. 6246--6252.

\bibitem{Hill2011}
T.~Hill and J.~Miller, ``How to combine independent data sets for the same
  quantity,'' \emph{Chaos: An Interdisciplinary Journal of Nonlinear Science},
  vol.~21, no.~3, pp. 033\,102 (1--8), 2011.

\bibitem{10.1007/978-3-319-93417-4_38}
M.~Schlichtkrull, T.~Kipf, P.~Bloem, R.~Berg, I.~Titov, and M.~Welling,
  ``Modeling relational data with graph convolutional networks,'' in
  \emph{Extended Semantic Web Conference}, 06 2018, pp. 593--607.

\bibitem{schulmanPPO}
J.~Schulman, F.~Wolski, P.~Dhariwal, A.~Radford, and O.~Klimov, ``Proximal
  policy optimization algorithms,'' \emph{ArXiv}, vol. abs/1707.06347, 2017.

\bibitem{liang2018rllib}
E.~Liang \emph{et~al.}, ``{RLlib}: Abstractions for distributed reinforcement
  learning,'' in \emph{Int. Conf. on Mach. Lear.}, 2018.

\bibitem{Haarnoja17}
T.~Haarnoja, H.~Tang, P.~Abbeel, and S.~Levine, ``Reinforcement learning with
  deep energy-based policies,'' in \emph{Proceedings of the 34th International
  Conference on Machine Learning (ICML)}, 2017.

\bibitem{Zhu2019}
H.~Zhu and J.~Alonso-Mora, ``Chance-constrained collision avoidance for mavs in
  dynamic environments,'' \emph{IEEE Robotics and Automation Letters}, vol.~4,
  no.~2, pp. 776--783, 2019.

\bibitem{Engstrom2020Implementation}
L.~Engstrom \emph{et~al.}, ``Implementation matters in deep rl: A case study on
  ppo and trpo,'' in \emph{Int. Conf. on Learning Representations}, 2020.

\bibitem{andrychowicz2021what}
M.~Andrychowicz \emph{et~al.}, ``What matters for on-policy deep actor-critic
  methods? a large-scale study,'' in \emph{International Conference on Learning
  Representations}, 2021.

\bibitem{domainRandomization}
J.~Tobin, R.~Fong, A.~Ray, J.~Schneider, W.~Zaremba, and P.~Abbeel, ``Domain
  randomization for transferring deep neural networks from simulation to the
  real world,'' in \emph{IEEE/RSJ Int. Conf. on Intelligent Robots and
  Systems}, 2017, pp. 23--30.

\bibitem{socialforces}
D.~Helbing and P.~Molnar, ``Social force model for pedestrian dynamics,''
  \emph{Physical Review E}, vol.~51, 05 1998.

\bibitem{airsim}
S.~Shah, D.~Dey, C.~Lovett, and A.~Kapoor, ``Airsim: High-fidelity visual and
  physical simulation for autonomous vehicles,'' in \emph{Field and service
  robotics}.\hskip 1em plus 0.5em minus 0.4em\relax Springer, 2018, pp.
  621--635.

\bibitem{makehuman}
``{Makehuman community.}, 2022.'' \url{http://www.makehumancommunity.org.}

\bibitem{yolov3}
J.~Redmon and A.~Farhadi, ``Yolov3: An incremental improvement,'' \emph{CoRR},
  vol. abs/1804.02767, 2018.

\bibitem{pythonlessons}
``{TensorFlow 2 YOLOv3 Mnist detection training tutorial.} pylessons, 2020.''
  \url{https://pylessons.com/YOLOv3-TF2-mnist.}

\end{thebibliography}

\end{document}